\documentclass{article}
\usepackage[utf8]{inputenc}
\usepackage{moreverb,url}

\usepackage[colorlinks,bookmarksopen,bookmarksnumbered,citecolor=red,urlcolor=red]{hyperref}
\usepackage{placeins}
\usepackage{csquotes}
\usepackage{graphicx}
\usepackage{float}
\usepackage{amsmath}
\usepackage{enumerate}

\usepackage[english]{babel}
\usepackage[T1]{fontenc}

\usepackage{authblk}

\usepackage{booktabs} 

\newcommand\BibTeX{{\rmfamily B\kern-.05em \textsc{i\kern-.025em b}\kern-.08em
T\kern-.1667em\lower.7ex\hbox{E}\kern-.125emX}}

\setkeys{Gin}{width=\columnwidth}

\begin{document}


\title{What's in the laundromat? Mapping and characterising offshore owned domestic property in London}

\author[1]{Jonathan Bourne \thanks{corresponding author: jonathan.bourne.15@ucl.ac.uk}}
\author[2]{Andrea Ingianni}
\author[2]{Rex McKenzie}

\affil[1]{University College London, The Bartlett Faculty of the Built Environment}

\affil[2]{Kingston University Department of Economics}


\maketitle

\begin{abstract}

The UK, particularly London, is a global hub for money laundering, a significant portion of which uses domestic property. However, understanding the distribution and characteristics of offshore domestic property in the UK is challenging due to data availability. This paper attempts to remedy that situation by enhancing a publicly available dataset of UK property owned by offshore companies. We create a data processing pipeline which draws on several datasets and machine learning techniques to create a parsed set of addresses classified into six use classes. The enhanced dataset contains 138,000 properties 44,000 more than the original dataset. The majority are domestic (95k), with a disproportionate amount of those in London (42k). The average offshore domestic property in London is worth 1.33 million GBP collectively this amounts to approximately 56 Billion GBP. We perform an in-depth analysis of the offshore domestic property in London, comparing the price, distribution and entropy/concentration with Airbnb property, low-use/empty property and conventional domestic property. We estimate that the total amount of offshore, low-use and airbnb property in London is between 144,000 and 164,000 and that they are collectively worth between 145-174 billion GBP. Furthermore, offshore domestic property is more expensive and has higher entropy/concentration than all other property types. In addition, we identify two different types of offshore property, nested and individual, which have different price and distribution characteristics. Finally, we release the enhanced offshore property dataset, the complete low-use London dataset and the pipeline for creating the enhanced dataset to reduce the barriers to studying this topic.

\end{abstract}

\section{Introduction}

The UK, and in particular London, is hub for international money laundering using property \cite{tiuk_stats_2022}. The volumes of money laundered through the UK capital are believed to be so high the city is sometimes refereed to as `The London Laundromat'. A substantial part of the reason for this is that in the UK it is possible for offshore shell companies to purchase property. This allows those with illicit incomes to buy property without being identified \cite{cowdock_cost_2018}. In the `National risk assessment of money laundering and terrorist financing 2020' The risk of money laundering using property purchases was rated as high \cite{hm_treasury_national_2020}, with domestic property being a particular focus due to the large amounts of money that can be moved in a single transaction and the lifestyle benefits such properties bring. In addition transparency international have found that 75\% of money laundering cases involving property use offshore shell companies \cite{de_simone_corruption_2015}. However, attempts to control money laundering have struggled for decades \cite{mitchell_accountants_1998}, for an in depth discussion of money laundering and property see \cite{tiwari_review_2020, unger_money_2011} In 2015 a list of offshore owned properties in England and Wales was leaked to, and made public by, the `Private Eye' a British publication \cite{private_eye_selling_2015}. Since then the UK government has published a regularly updated dataset of offshore owned properties called the `Overseas companies that own property in England and Wales - Use land and property data' (OCOD) \cite{land_registry_overseas_2022}.

The OCOD dataset has the potential to be a great resource for those wishing to understand a the distribution and characteristics of offshore owned property in the UK. However, although the dataset has been used by researchers \cite{mckenzie_anchoring_2020, sa_effect_2016}, the dataset is challenging to use with addresses recorded as free text. These addresses can be incomplete, contain nested properties, where multiple properties exists within a single row or title number, it also contains no information on whether the property is domestic, business or something else. Such poor quality data makes understanding the distribution and characteristics of offshore owned property in the UK challenging.

In addition to offshore owned domestic property, London has large amounts of short term lets and low-use/empty property. In this paper the Offshore domestic property, airbnb property and low-use property, will be grouped under the term `Unconventional Domestic Property' (UDP). Short term lets of the sort facilitated by Airbnb, One Fine Stay and Vrbo have recieved a quantiative exploration In both London and other major cities \cite{shabrina_airbnb_2022, bao_impact_2020, garcia-lopez_short-term_2020}. Low-use property, i.e. property with no permanent resident, has also received coverage \cite{bourne_empty_2019, noh_urban_2021, baba_factors_2019}, but less than the short term lets market due to difficulty obtaining data which is seldom publicly available due to privacy concerns. In the UK data gathering is further complicated by the distinct sub-national jurisdictions. For example, the register of property sales is public in England and Wales \cite{land_registry_how_2022} but not in Scotland. To counter some of the difficulties with property data the UK government developed the Unique Property Reference Number (UPRN) \cite{os_os_2021}. The UPRN is a referencing system designed increase the ease of linking property data across government datasets. However, whilst the use of the UPRN is mandated, almost no government department uses it meaning linking the data requires advanced data processing skills \cite{chi_new_2021, comber_machine_2019}.

A partial solution to the data issues within OCOD and other property datasets, is the use of address parsers. Recently opensource address parsers have been developed to convert free text addresses to a tabular and normalised format, which is much easier to analyse \cite{yassine_leveraging_2020, barratine_libpostal_2017, delil_parsing_2020}. 
Address parsing is a form of Named Entity Recognition (NER) which is a subfield of Natural Language Processing (NLP) which is itself a subfield of Machine Learning (ML). NER uses statistical techniques to identify certain words within a sentence or larger block of text  which represent a predefined entity and labels them with a given entity class. For address data,  an example of a specific class could be `street name' or `postcode'. The named entity is defined by its `Span' which is the position number of the start and end characters as well as the label of the entity class. 

Although very valuable the existing address parser's are not without problems. Many of the parsers are are trained using Open Street map data \cite{osm_openstreetmap_2004}, this has the advantage that there are many millions of stored addresses to use as training data, and algorithms can be trained to parse addresses from countries across the world using multiple languages and scripts. However, it also means that even though training data has noise induced to increase robustness, competing algorithms are all trained on the same dataset. This training dataset may be subtly different from the actual data the resulting algorithms are applied to. This can lead to weaknesses, for example the most well established parser libpostal \cite{barratine_libpostal_2017} has been shown to have issues parsing incomplete addresses \cite{yassine_leveraging_2020}, it also struggles with addresses in formats it is not used to \cite{craig_scaling_2019}. A more fundemental issue may be that the algorithms are designed on the assumption that they are parsing a single address, the OCOD dataset contains large numbers of nested addresses where a single free text line may contain tens or even hundreds of properties. As such a custom parser must be created to deal with the issues presented by this specific dataset. Given that there is no real training data creating such a parser will need to take advantage of recent developments in ML and NLP.

A major difficulty in any machine learning project is the obtaining of labelled data. This is because the majority of machine learning tasks use supervised learning. In supervised learning, the algorithm is instructed to learn a specific task such as classification or regression, for which the training data supplies the correct answer, the algorithm is considered trained when the error between the predictions and ground truth is minimised. The data used in supervised classification tasks typically has to be labelled by a human. This is time consuming, and can be impractical for small teams or those without funding to pay others using `crowd sourcing' \cite{sheng_machine_2019, zhang_learning_2016}. A developing solution to this issue is `Weak learning' \cite{ratner_data_2017}, in this approach data is labelled using simple rules, this can produce a large number of labels which are often correct, but can also overlap and contradict one-another. A denoising process is then applied to the data using some statistical method such as a hidden markov model \cite{lison_named_2020} or matrix completion \cite{ratner_training_2018}. The denoising process removes overlapping labels to minimise the error with the ground truth. An obvious critique of weak learning is that the true labels are not known bringing prediction quality into question, however one of the most powerful aspects of weak learning is that it can produce models that are almost equal in performance to hand labelled data \cite{fries_ontology-driven_2021}, indeed the quality of the model scales according the the size of the unlabelled dataset \cite{ratner_data_2017}. In addition recent research has highlighted that most datasets have errors even those used as a benchmark machine learning datasets such as MNIST, ImageNet and Amazon reviews \cite{northcutt_pervasive_2021}; this suggests that datasets created either through weak learning or hand labelling are on an error scale but can both produce high quality models.

This paper proposes a new method of identifying and classifying offshore owned property in England and Wales and uses this method to answer the research question; How is offshore domestic property in London characterised in terms of financial value and spatial distribution? How does this differ from low-use property, airbnb property and conventional homes? Does offshore domestic property appear to be broadly homogeneous, or does it contain distinct sub-groups? 

\section{Method} 

The method is broken into three parts. The first part introduces the data, the second part details how the OCOD dataset is cleaned and transformed into the enhanced OCOD dataset. The third part describes the steps used in the analysis. 

It should be noted that although address parsing is a major part of this project, the goal is not to create an universal address parser. As such the focus of the method is the process of creating the enhanced OCOD dataset and the resulting analysis. 

In this project geographical analysis is performed on discrete units. These units were created by the Office for National Statistics (ONS) for use in census geography \cite{ons_census_2022}. We use 3 levels of census geography which are (from smallest to largest) the Output Area (OA), Lower Super Output Area (LSOA) and Middle Super Output Area (MSOA), In addition we use the Local Authority District (LAD), which is the smallest level of local government. On average a LAD contains 30 MSOA although there can be substantial variance. The use of the OA system provides a standard interface with government data and prevents data privacy issues. 

\subsection{Data and code}

This work uses several publicly available datasets. They key data set is the `Overseas companies that own property in England and Wales - Use land and property data' (OCOD) \cite{land_registry_overseas_2022}. The data cleaning process uses the Land Registry Price Paid dataset \cite{land_registry_how_2022}, The VOA ratings listing dataset \cite{voa_voa_2017} and the Office of National Statistics Postcode Directory (ONSPD) \cite{ons_onspd_2022}. The Airbnb data is from the website InsideAirbnb \cite{cox_inside_2022}, The airbnb properties in this dataset only include entire homes which are let to the extent it is not possible to have a permenent tennent. The low-use dataset uses data gathered using Freedom of information (FOI) requests, the majority of which was collected for \cite{bourne_empty_2019}, but the outstanding London boroughs have since been added using the same FOI process (see github repo \href{https://github.com/JonnoB/lsoa_foi_form_maker}{lsoa foi template maker}). The definition of low-use is the same as \cite{bourne_empty_2019} which is is broadly that the property has no permanent resident, but this is through choice. Price data used in the inference and analysis section comes from the Price Paid dataset. The London Low-use dataset and the Enhanced Offshore ownership dataset are both available along with the code used at the github repository for this project  \href{https://github.com/JonnoB/empty_homes_london}{JonnoB empty homes london}. The data description of the Enhanced OCOD can be found in supplementary material.

It should be noted that the nested properties highlight another nuance with the OCOD dataset. Nested properties are part of a title that is owned by an offshore company but that title may refer to the freehold as apposed to the lease hold. This means that the right to live in the property has been sold to an individual for some period of time (e.g. 100 years) although the property still belongs to the freeholder. Leaseholders usually pay `ground rent' to the freeholder. This leaseholds are typically case with flats and apartments. We included both freehold and leasehold properties on the basis that in either case the benefit either in terms of sale or 'ground rent' asset value belongs to the offshore company.

\subsection{Creating the Enhanced OCOD dataset}

\begin{figure}
    \centering
    \includegraphics{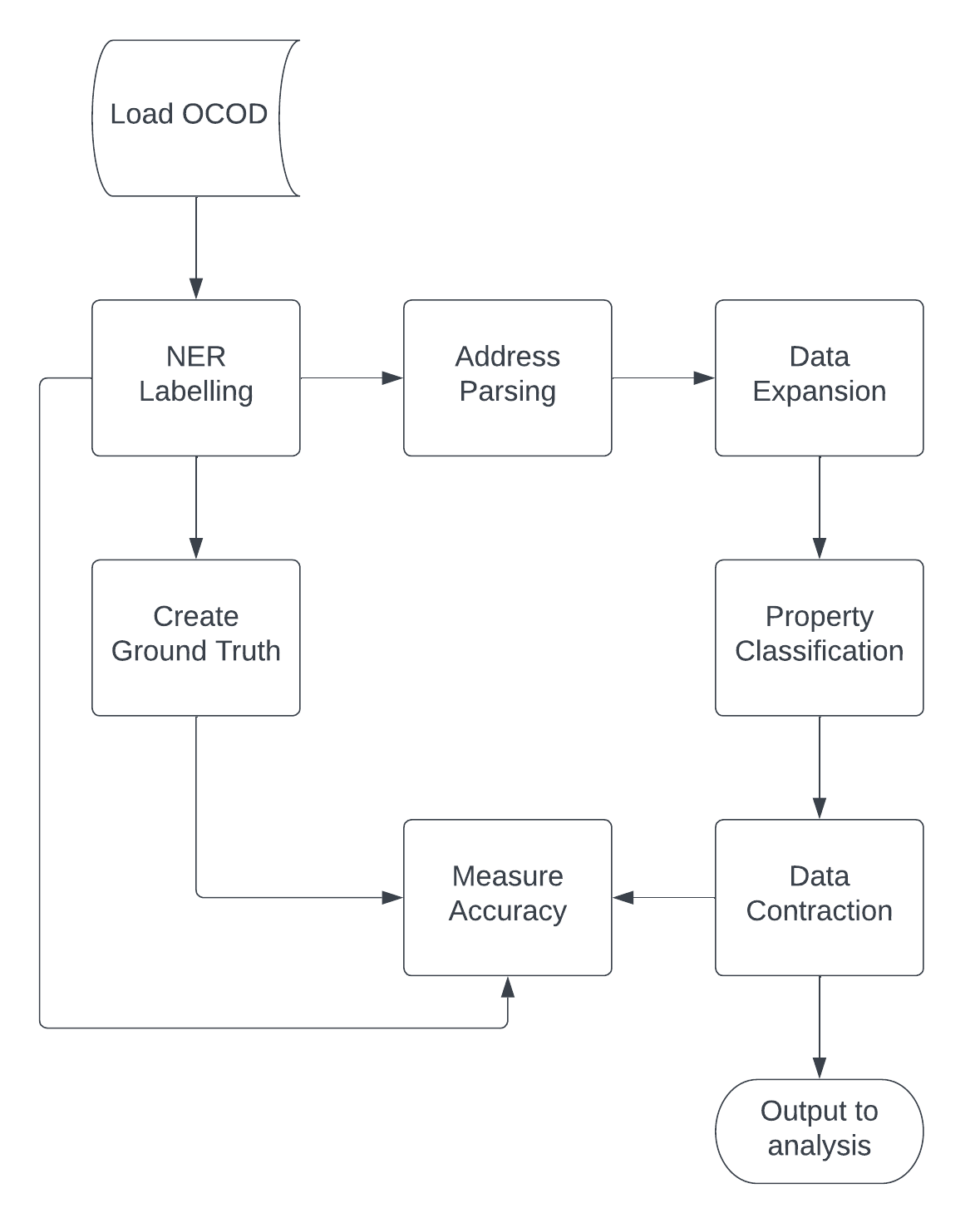}
    \caption{A schematic of the data cleaning process, from loading the OCOD dataset to ouputting the final dataset ready for analysis}
    \label{fig:cleaning_process}
\end{figure}

The enhanced OCOD dataset is created by putting the original OCOD dataset through a process of cleaning, labelling, parsing and classification.
The end result is a dataset that can be easily analysed and follows tidy data principles \cite{wickham_tidy_2014} in that each property is an observation/row in the data. The OCOD dataset is a CSV file, it contains several columns however for the purpose of this paper two are of importance, the title number and the property address. The title number is the unique reference given to each `property' although in reality a title can contain many properties or none. The property address contains the free text which will be parsed.
The data cleaning and preparation process is broken into five sections, Label, Parse, Expand, Classify and Contract, in addition quality testing of the labelling and classification is performed. A schematic of the process is shown in figure \ref{fig:cleaning_process}.

The address data has a form similar to the three examples below. As the data is all addresses there is a degree of structure and predictability to it. The addresses are broken into comma separated parts; e.g. if present, street number or unit ID is always first, and postcode is always last, (see table \ref{tab:address_examples}).

\begin{table}[ht]
\begin{tabular}{|c|l|}
\hline
Example & Address                                             \\ \hline
1       & flat 6, chartfield house, babel road, london        \\ \hline
2       & 5 to 15 (odds only) babel road, london (w1 8ap)     \\ \hline
3       & 5 babel road, london and parking 3.5 w1 8ap \\ \hline
\end{tabular}
    \caption{Example address formats}
    \label{tab:address_examples}
\end{table}

After loading the OCOD dataset light pre-processing was performed such as ensuring a space after comma and removing double spaces, changing to lower case etc. The data was then imported to programmatic \cite{humanloop_humanloop_2022} where entities were labelled using regex (Regular expressions) functions. There were around 50 functions used to describe the 8 types of named entity (Regex functions can be found here  \url{https://github.com/JonnoB/empty_homes_london/tree/main/humanloop_1_2/export}). These entities are unit ID, unit type, building name, street number, street name, city, postcode, filter type. The entity classes are described in table \ref{tab:entity_classes}.

\begin{table}[ht]
    \begin{tabular}{|p{0.35\linewidth} | p{0.6\linewidth}|}
\hline
Entity class  & Description                                                                                                                                                                                                                                                                  \\
\hline 
Unit ID       & Describes a sub-unit such as a flat number or parking space ID. Example one would have `6' and example three would have `3.5' as unit id. Unit Id is not always a number                                                                                                     \\
Unit type     & Describes the type of unit if available. Example one would have `flat' whilst example three would have "parking"                                                                                                                                                       \\
Building Name & Example one would have `chartfield house', the field would not be present for the other two examples                                                                                                                                                                         \\
Street Number & The street number of the property if available, would be `5 to 15' in example two and `5' in example three. Street number is not always a number                                                                                                                             \\
Street Name   & Self explanatory, would be `babel road' in all three examples                                                                                                                                                                                                                \\
Filter Type   & When multiple properties are included in the address a filtering condition is often used, this is because in the UK odd and even numbers are often on opposite sides of the road. Or a company may not own all the flats in an apartment block. Example two would have `odd' \\
City          & Self explanatory, would be London for all three examples                                                                                                                                                                                                                     \\
Postcode      & Self explanatory. In almost all cases the post code is in parenthesis. In addition UK postcodes follow a pattern which can be extracted using regex, making them easy to label  
\\ \hline

    \end{tabular}
    \caption{Definitions of the entity classes used in the labelling process, example formats are found in table  \ref{tab:address_examples}.}
    \label{tab:entity_classes}
\end{table}

Once the labels have been added to the data the whole dataset is extracted as a json file. In some cases label spans overlap these are removed either using a simple rule that keeps only the largest span, or a denoising process that uses a hidden markov model, following the principles of weak labelling. In addition, using the output of Programmatic, we create a predictive model using spaCy \cite{honnibal_spacy_2017}, the model uses RoBERTa \cite{liu_roberta_2019} as the underlying algorithm. RoBERTa is a type of transformer \cite{vaswani_attention_2017}, currently one of the most successful approaches to solving problems in NLP. 

Once the labels have been denoised a ground truth data set is hand labelled using 1000 randomly sampled observations. An additional dev set of 2000 examples is also labelled using the same process. This ground truth dataset is used to asses the quality of the NER labelling, using programmatic and spaCy.

\subsubsection{Address parsing}

With a single address the parsing process is effectively over when the the entities have been labelled. However, in the case when there are multiple properties the parsing process is more involved the label dependency becomes important. As an example consider the nested address 'flat 5, chartfield house and flat 16, zebra house, babel road, london (w1 8ap)'. In this example there are two flats in two separate buildings on a single street and with a common postcode, as such there are two distinct addresses with some shared labels. We parsed the addresses by assigning each character span its own row and each label class its own column. We then propergated the columns backwards to create complete address rows. Certain labels were classed as terminators and all non-terminator rows were removed from the data. Terminators were hierarchically structured such that if the most high priority label class was not available the next label class in the hierarchy was used. The terminator list in hierarchical order is unit type, unit ID's, building name, street number. The presence of a span in a label column would block the backfilling of data from later in the sentence, this prevented situations where houses would be assigned the wrong road. When parsing was complete and address rows without terminators were removed and a tabular dataset of parsed addresses was returned.

As mentioned earlier and shown in table \ref{tab:address_examples}, some title numbers contained multiple addresses using an abbreviated format. To have a single address per line required that the dataset be expanded.

\subsubsection{Address Expansion}

Expanding the dataset so that each row represents a single property required machine reading of the unit ID label and the filter number label. A regex expression was used to identify unit ID's or street numbers that covered multiple addresses the first and last numbers were then extracted and the numbers between them filled in making a contiguous list of numbers from i to j. If a filter number label was present it was used to identify whether the numbers were odd or even and the list of numbers were filtered to so only the appropriate number type remained. With the data now parsed and expanded such that a single property address was represented on a single line the data could now be localised to a specific OA or LSOA and classified into a specific property type. 

\subsubsection{Address Classification}

\begin{figure}
    \centering
    \includegraphics{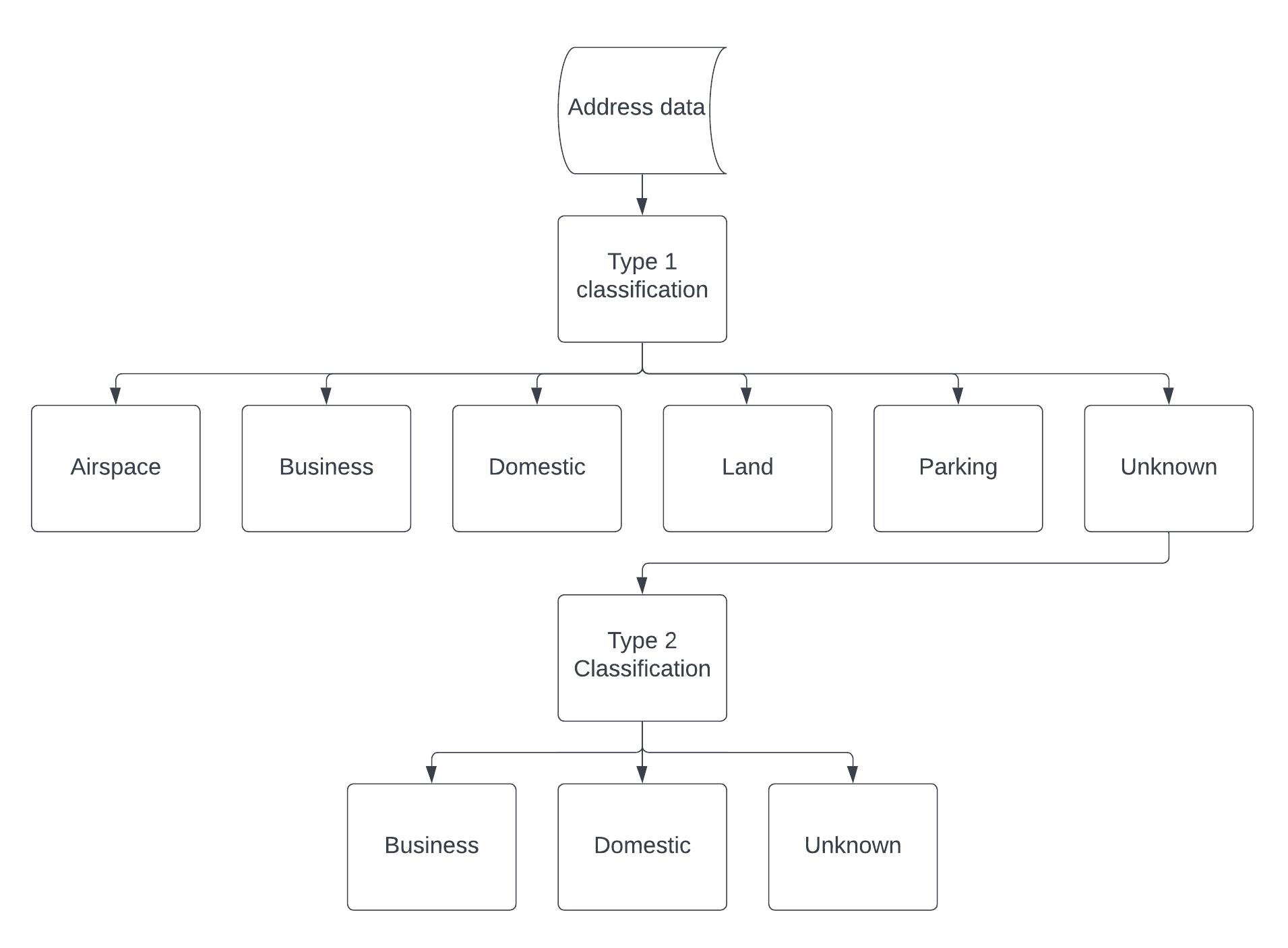}
    \caption{The classification process during the preparation of the enhanced OCOD dataset.}
    \label{fig:classification_process}
\end{figure}

The classification phase of the cleaning process includes classifying the property types but also includes localising the properties within the UK governments geographical area system. Initially all postcodes are cross referenced with the Office of national Statistics' Postcode Database (ONSPD). This links the postcode and the OA system used by the ONS for census and other demographic analysis. However, not all addresses have a postcode, in these cases the parsed addresses are cross referenced with the land registry price paid dataset and the VOA company ratings list dataset, if a match is found the OA and LSOA is added to the OCOD address. For nested addresses the same OA and LSOA is added to all addresses within the property title. The data is classified into one of six classes, in a sequential process (see figure \ref{fig:classification_process}). The 6 classification groups are airspace, business, domestic, land, parking and `unknown'. There are two classification processes Type 1 which has twelve classification steps, where each step identifies some aspect of a single property class, (for details on the process see  \href{https://github.com/JonnoB/empty_homes_london/blob/main/locating_and_classifying_the_ocod_dataset.ipynb}{classifying\_the\_ocod\_dataset.ipynb}, section `Classify Property type'). Any properties which cannot be identified are classed as `unknown', these properties are almost entirely a mix of domestic and business. Once the Type 1 classification process is complete the Type two process is applied. As shown in Figure \ref{fig:classification_process} the Type two process builds on the output of Type 1 and deduces whether a property is domestic or business by finding which addresses can be identified. Type 2 has several sequential classification steps which filter for the level of completeness of an address. When class deduction is possible the Type two process changes the `unknown' class of the Type one process to `domestic' or `business' as appropriate. However addresses which do not contain enough information to accurately locate remain in the `unknown' class. The outputs of the processes are provided separately in the dataset as the Type 1 process is more secure as it primarily relies on positively identifying classes whereas Type two uses deduction to work out what the property is not. The analysis will use Type two classification.

\subsubsection{Address Contraction}

This is the final and most straight forward of all the cleaning steps. As it is not unusual for commercial premises to occupy more than one `street address' non-domestic properties that are within a nested property group are reduced back to single address. As such only domestic properties are ever within a nested property title. 

\subsubsection{Test set and Accuracy metrics}

Whilst the previous sections have described the steps by which the data is prepared, they do not describe any way to measure the quality of the labelling or classification processes. This is essential as the for the final analysis to be considered reasonable the data needs to be of high quality. The data is tested by randomly sampling 1000 addresses from the original dataset these samples are then hand labelled and held out from any further analysis until the end of the cleaning process. Figure \ref{fig:cleaning_process} shows that the ground truth is created after the initial labelling, this is for practical reasons to speed up the hand labelling process as many of the labels will already be complete and correct. Finally the classification quality is assessed. The manual checking of the class is done by looking up the address using map software, looking at pictures of the property, and doing an internet search of the address to see if it is a domestic or a commercial property. This process is certainly not infallible but without the use of the UPRN system there is no other way within a reasonable time frame.

Once the ground truth has been created, the datasets produced by the processes in this paper and alternative parsing algorithms can be compared. Performance will be measured using the precision, recall and F1 metrics. Precision is the probability of a span exactly matching the ground truth test set. In the binary class case precision is defined as \begin{equation}
    \mathrm{Precision}= \frac{tp}{tp +fp},
\end{equation} where $tp$ is the total number of true positives where the span matches the test set and $fp$ is total number of false positives where span does not. Recall is the fraction of matched spans which are also given the correct class, it is defined by \begin{equation}
    \mathrm{Recall} = \frac{tp}{tp +fn},
\end{equation} where $fn$ is the total number of false negatives. The F1 score is \begin{equation}
    \mathrm{F1} = \frac{2tp}{2tp + fp + fn},
\end{equation} making it the harmonic mean of precision and recall. However in order to adjust the measure for the multiclass data used in this paper we use the micro-averaged values. Microaveraging takes the sum of $tp,fp,fn$ across all classes, and uses those summed values to calculate the final precisions, recall, F1 values. As an example micro precision can be defined as $\bar{tp}_\mu =\frac{\sum^m_i tp_i}{\sum^m_i tp_i + \sum^m_i pp_i}$, where $i$ is the property class and $m$ is the total number of properties. It should be noted that although the micro average can produced biased results in imbalenced data, the output put is more easily interpretable as it shows in simple terms the fraction labels that are correct.

Once the final enhanced dataset has been created the process of analysis can begin. 

\subsection{Analysing the offshore dataset}

The analysis of this paper is performed on each domestic property class. Where the domestic property classes are conventional homes, low-use homes, airbnb properties, offshore homes. The purpose of the analysis is to understand the differences in the behaviour between these property classes and the implications for the housing market. Three core analyses are performed in this paper, inferring property price, entropy of property and spatial autocorrelation. Having described the outline of the method the next section describes how the total number of UDPs will be calculated.

\subsubsection{Calculating the total number of unconventional domestic property}
The three property types airbnb, low-use that make up UDP are not mutually exclusive and so cannot simply be added together. Instead the joint probability will be found under two conditions; The first is that data are independent; The second that offshore property is always low-use. The first case is the null hypothesis and will act as the upper bound, whilst the second case is based on offshore properties being inherently luxury purchases and the owner having multiple homes. The second case reduces the total number of classes as the offshore properties become a subset of the low-use properties. The probability of a UDP in a given LSOA is defined as \begin{equation}
    P(U|m)=1-\prod_{i=1}^m (1-P(i|m)),
\end{equation} That is 1 minus the joint probability of a property not being any UDP type. Although the number of properties will vary if sampled from the distribution the variance will be small compared to the difference between the two cases where UDP types are independent or not. Therefore for simplicity the total number of UDPs in an LSOA is calculated to be the product of the total number of homes and the probability of being an UDP in that LSOA. With the number of UDPs calculated the next section will describe the sampling process applied to the number of UDPs to infer their mean prices.

\subsubsection{Inferring the prices of properties in London}
\label{sect:method_sample}

In \cite{bourne_empty_2019} The value of domestic property and low use property was inferred using a simple graphical model. The same technique will be used in this work to infer the value of the distinct UDP types as well as for the total number of UDPs. Due to low the number of samples in certain areas, the LSOA counts will be aggregated up to MSOA level.

The distribution of the mean price by property type is calculated as shown in \begin{equation}
        \left \langle C \right \rangle _{m = i} = \frac{1}{z}\sum_j^z C_j P(C_j \mid m = i),
    \label{eq:distribMean}
\end{equation} Where $\left \langle C \right \rangle _{m = i}$ is the mean price for property type $m$, $C_j$ is the vector of possible prices for $j$th property and $P(C_j \mid m = i)$ is the probability price distribution for $C_j$. The sum across all values of $j$ is divided by the total number of properties $z$. Clearly $\left \langle C_j \right \rangle _{m = i}$ is simply the mean price of the MSOA of $j$ as such equation \ref{eq:distribMean} is the Bayesian equivalent mean of the mean MSOA price weighted by distribution of homes. Both methods can be used and for large values will produce similar results, however equation \ref{eq:distribMean} is able to provide an indication of the variance statistical significance and so will be used here. The equation is implemented as form of ancestral sampling from the distribution, in order to obtain a representative mean price, the sampling is repeated 501 times to produce a distribution of the mean, this is similar to the bootstrap technique developed by \cite{efron_introduction_1993}.

\subsubsection{Analysing the entropy of domestic property}

In information theory Shannon entropy \cite{shannon_mathematical_1948}  is used to describe the amount of disorder within a system of a discrete fixed number of states and is measured in the amount of bits of information required to store it. The higher the number of bits the more disorder there is in a system and conversely the lower the number of bits the more order there is in a system and the easier it is to describe. The equation for Shannon entropy is \begin{equation}
    E = -\sum_{i=1}^{i=k}p_i \mathrm{log}p_i,
\end{equation} where $k$ is total number of states, and $p_i$ is the probability of the $i$th state.

In the geospatial analysis being performed here the entropy describes how concentrated the offshore property types are. The discrete states in this case are the 4835 LSOA in London. The entropy would be minimised if all of a certain property class would be found in a single LSOA whilst entropy would be maximised if the properties were spread evenly across all LSOA. As such it can be said that entropy represents how concentrated the property classes are. 

Although understanding the concentration of the property is very valuable it lacks any spatial information, for this we need to use a measure of spatial auto-correlation.

\subsubsection{Analysing the spatial auto-correlation of domestic property}

Spatial auto-correlation measures the strength of the correlation between objects in physical space and a given property. That is, if objects close together have very similar values but objects far away are very different, there will be a high auto-correlation value; whilst if objects that a close by have a very different value there will be a very low value-of auto-correlation. In this paper we will use Moran's I \cite{moran_notes_1950} as the measure for spatial auto-correlation. Moran's I is described as \begin{equation}
   I={\frac {k}{W}}{\frac {\sum _{i=1}^{k}\sum _{j=1}^{k}w_{ij}(x_{i}-{\bar {x}})(x_{j}-{\bar {x}})}{\sum _{i=1}^{k}(x_{i}-{\bar {x}})^{2}}},
\end{equation} where $k$ is the total number of spatial indexes in this case LSOA, $x$ is the target characteristic e.g. number of offshore properties, $\bar{x}$ the mean characteristic value, $W$ is the matrix of weights and $w_{i,j}$ is element $i$,$j$ of the weights matrix. Similar to the more commonly used, non-spatial, Pearson correlation, Moran's I varies between -1 and 1, where 1 shows perfect positive correlation. 

Having described both the data processing and the analysis methods, we will now introduce the results of both parts of this method.

\FloatBarrier

\section{Results}

The Results section is broken into two parts the first describing the data processing and the second the analysis of the enhanced OCOD dataset.

\subsection{Labelling and parsing performance}

Table \ref{tab:NER_fscore} shows the labelling results of the NER task. These results are similar for both the basic process and using a trained neural network. Overall the model scored an f1 of 0.96 although this was somewhat unevenly distributed across the categories. In order to accurately locate and class the addresses building name, street number, street name and unit id are required (postcode is already supplied), using just these metrics the mean f1 is 0.95.

Once the labelled addresses have been parsed and expanded the process of locating the properties and classifying them is performed. The f1 of the classification process is 0.94 across all classes and 0.96 for domestic properties with a precision of 0.93 and a recall of 0.99, as shown in table \ref{tab:classification_f}. This means that whilst almost all the domestic properties are being correctly tagged the process is slightly over-estimating the number of domestic properties and under estimating the total number of businesses. Counting only unique title numbers instead of properties reduces the f1 to 0.92 as a large number of nested domestic properties are discounted

\begin{table}[ht]
    \centering
\begin{tabular}{lrrrr}
\toprule
        class &  precision &  recall &  fscore &  support \\
\midrule
building name &       0.93 &    0.89 &    0.91 &    383.0 \\
         city &       1.00 &    0.97 &    0.98 &    947.0 \\
number filter &       1.00 &    1.00 &    1.00 &     23.0 \\
     postcode &       1.00 &    1.00 &    1.00 &    768.0 \\
  street name &       1.00 &    0.95 &    0.97 &   1029.0 \\
street number &       0.98 &    0.98 &    0.98 &    678.0 \\
      unit id &       0.98 &    0.92 &    0.95 &    370.0 \\
    unit type &       1.00 &    0.97 &    0.98 &    488.0 \\
\bottomrule
\end{tabular}

    \caption{The NER labelling task is uneven across the entity types but overall provides an f1 score of 0.95-0.96, depending on whether only core labels or all labels are considered.}
    \label{tab:NER_fscore}
\end{table}

\begin{table}[ht]
    \centering
\begin{tabular}{lrrrr}
\toprule
   class &  precision &  recall &  fscore &  support \\
\midrule
airspace &       1.00 &    0.93 &    0.96 &     14.0 \\
business &       0.97 &    0.78 &    0.87 &    311.0 \\
 carpark &       1.00 &    0.96 &    0.98 &     26.0 \\
domestic &       0.93 &    0.99 &    0.96 &    918.0 \\
    land &       1.00 &    0.99 &    1.00 &    179.0 \\
 unknown &       0.00 &    0.00 &    0.00 &      9.0 \\
\bottomrule
\end{tabular}
    \caption{The performance of the classification task shows that the most challenging type are domestic and business.}
    \label{tab:classification_f}
\end{table}

As the labelling and classification f1 scores show good quality results subsequent analysis of the data can be treated as having a reliable foundation. The next section will use the enhanced OCOD dataset to provide insight into UDPs in London. Overall the enhaced ocod dataset has 138 thousand entries 44\% more than the 94 thousand in the original dataset. Table \ref{tab:prop_type} provides the breakdown by property use of the enhanced OCOD dataset using the type 2 classification.

\begin{table}[ht]
\centering
\begin{tabular}{rlrr}
  \hline
 & class & counts & percentage \\ 
  \hline
1 & airspace & 1667 & 1 \\ 
  2 & business & 21396 & 16 \\ 
  3 & carpark & 2438 & 2 \\ 
  4 & domestic & 94576 & 69 \\ 
  5 & land & 16384 & 12 \\ 
  6 & unknown & 1124 & 1 \\ 
   \hline
\end{tabular}
\caption{Breakdown of property type using type 2 classification} 
\label{tab:prop_type}
\end{table}

\subsection{Data analysis}

\begin{figure}
    \centering
    \includegraphics{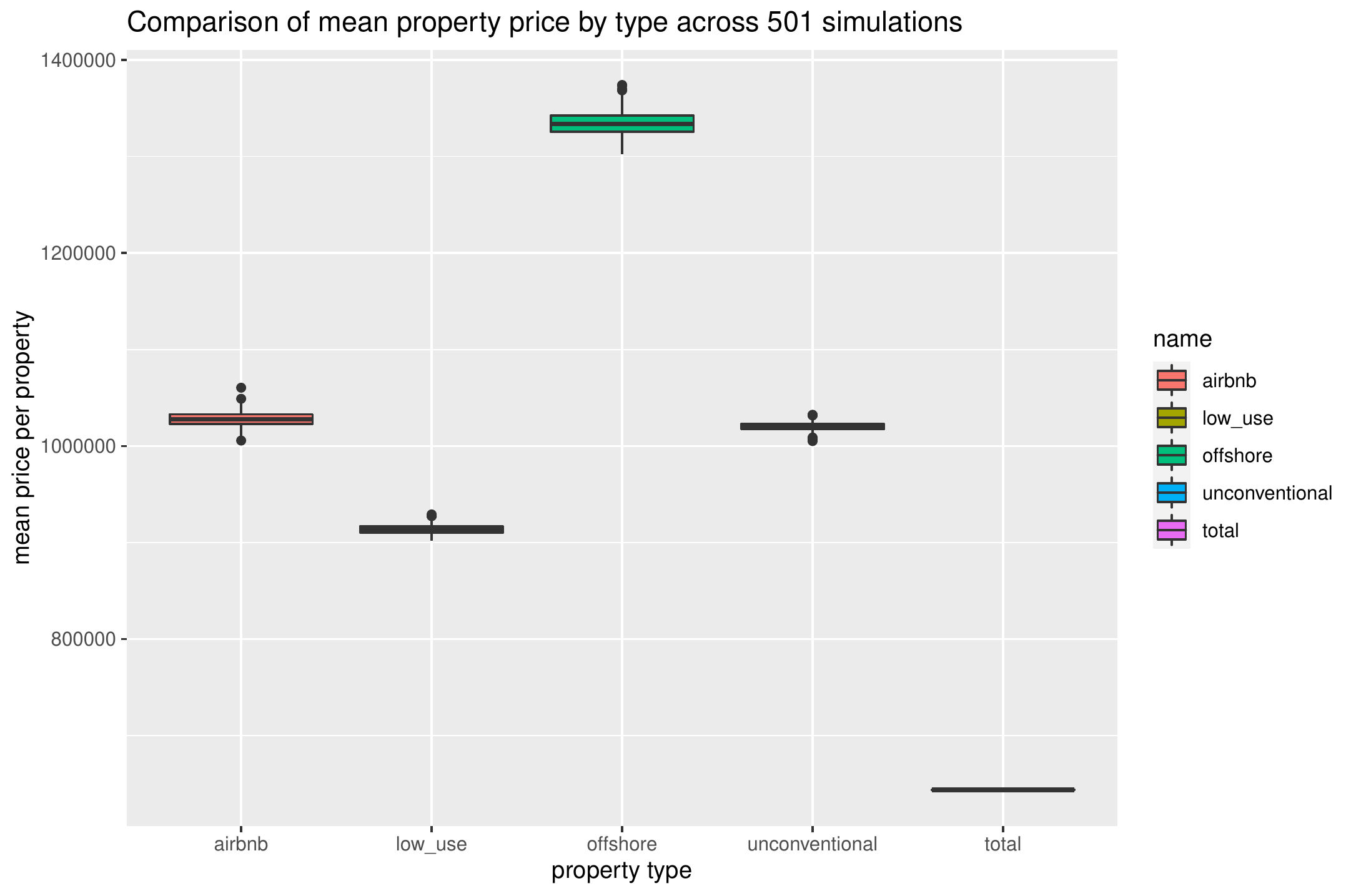}
    \caption{All three of the property types under analysis are worth susbtantially more than regular homes in the capital. Airbnb properties are worth more than low-use homes, highlighting the importance of location for this property type that exceeds that of general low-use property. However, offshore property is by far the most expensive of all the property types}
    \label{fig:mean_property_price}
\end{figure}

\begin{figure}
    \centering
    \includegraphics{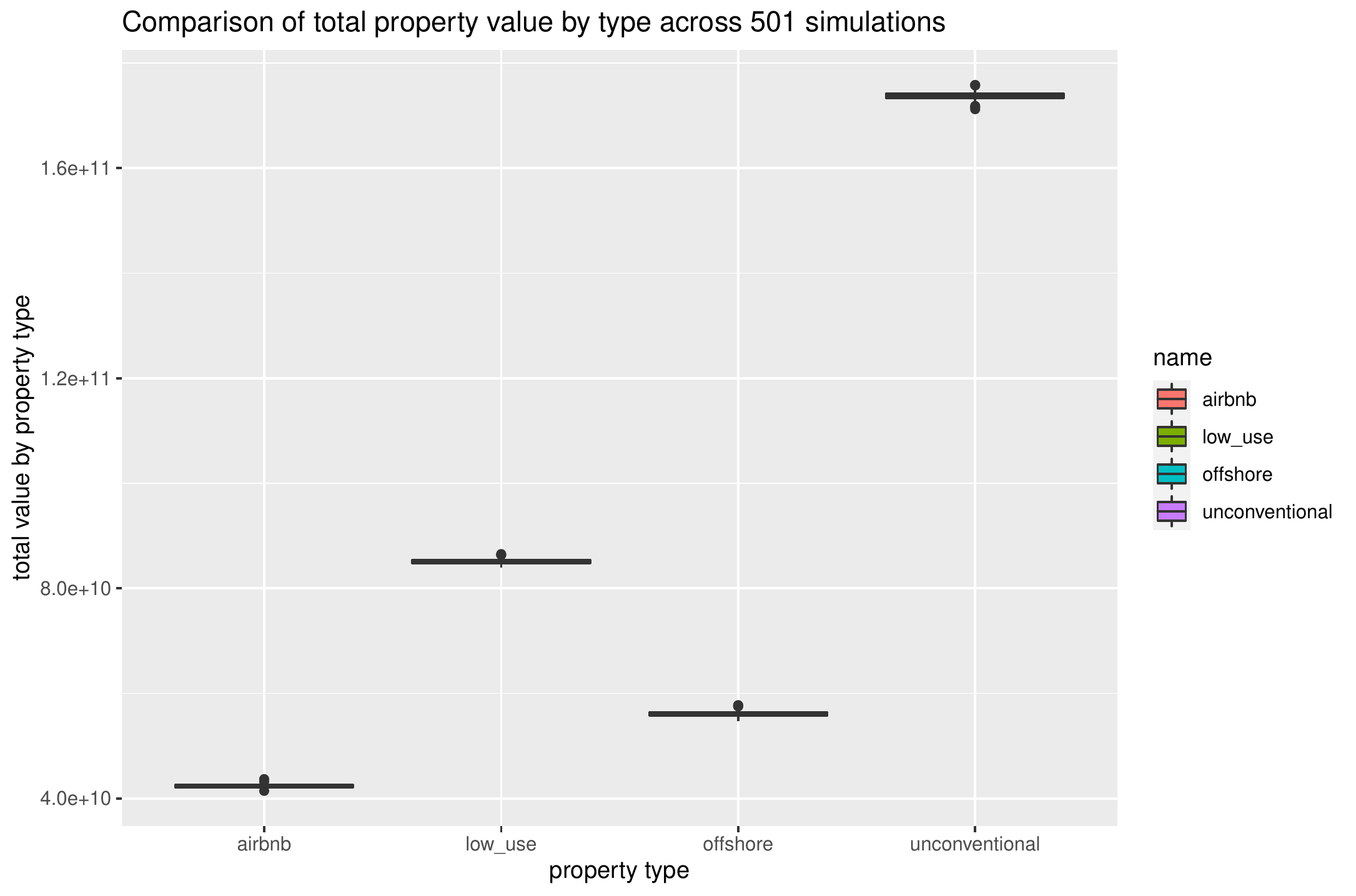}
    \caption{Due to the quantity of low use homes, their total value dwarfs that of the offshore domestic property, even though offshore domestic property is susbtantially more expensive}
    \label{fig:total_value}
\end{figure}

Analysing the subset of the enhanced OCOD dataset that is domestic property in London and comparing it to Airbnb, low-use and conventional property reveals  certain characteristics of these property types. A summary of the quantitive analysis can be seen in table \ref{tab:metrics_table}. Figure \ref{fig:total_value} shows the total value of the offshore properties is approximately \pounds 56 Billion, this is slightly than the total value of airbnb property of \pounds 42 Billion, however both are less than the value of the low-use property in London which is \pounds 85 Billion. Although the total amounts are interesting, it is more insightful to compare the mean value of the property types. There are significant differences in the mean values of the the property types as shown in Figure \ref{fig:mean_property_price}. Mean value of the Offshore property is \pounds 1.33 million, substantially higher than Airbnb properties which have a mean value of \pounds 1 million, this is followed by Low-use properties at \pounds 900 thousand. Overall the mean value of the UDPs is \pounds 1.02 million. However, all UDPs are much more expensive than the mean conventional property price of \pounds 600 thousand. 

\begin{figure}
    \centering
    \includegraphics{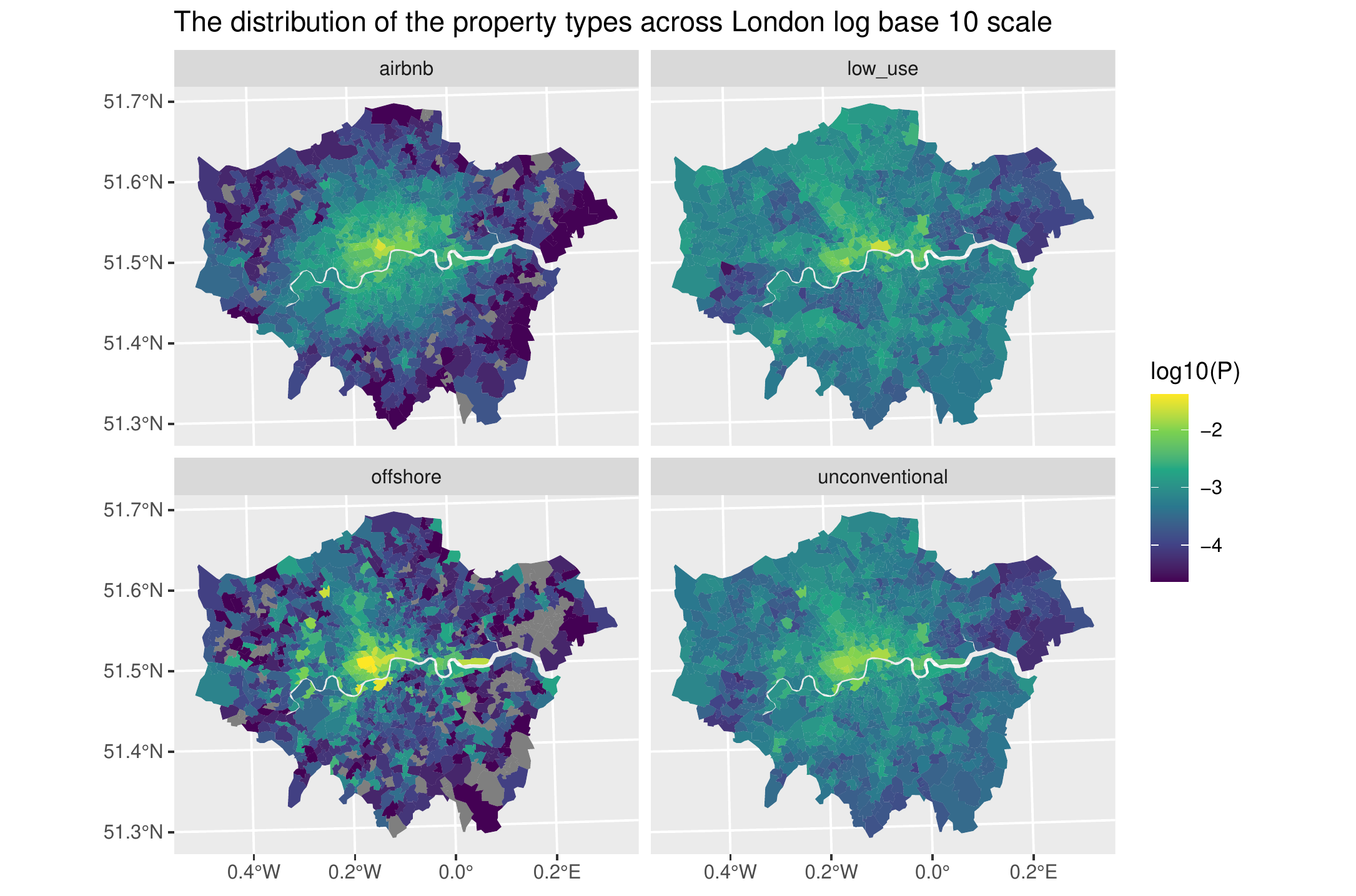}
    \caption{Looking at the maps clearly shows that there is a substantial difference in the distribution of the three property types. Offshore domestic property has some extremely high concentrations where an entire housing development is owned by an offshore company. Grey areas have no properties of that type present}
    \label{fig:property_maps}
\end{figure}

\begin{figure}
    \centering
    \includegraphics{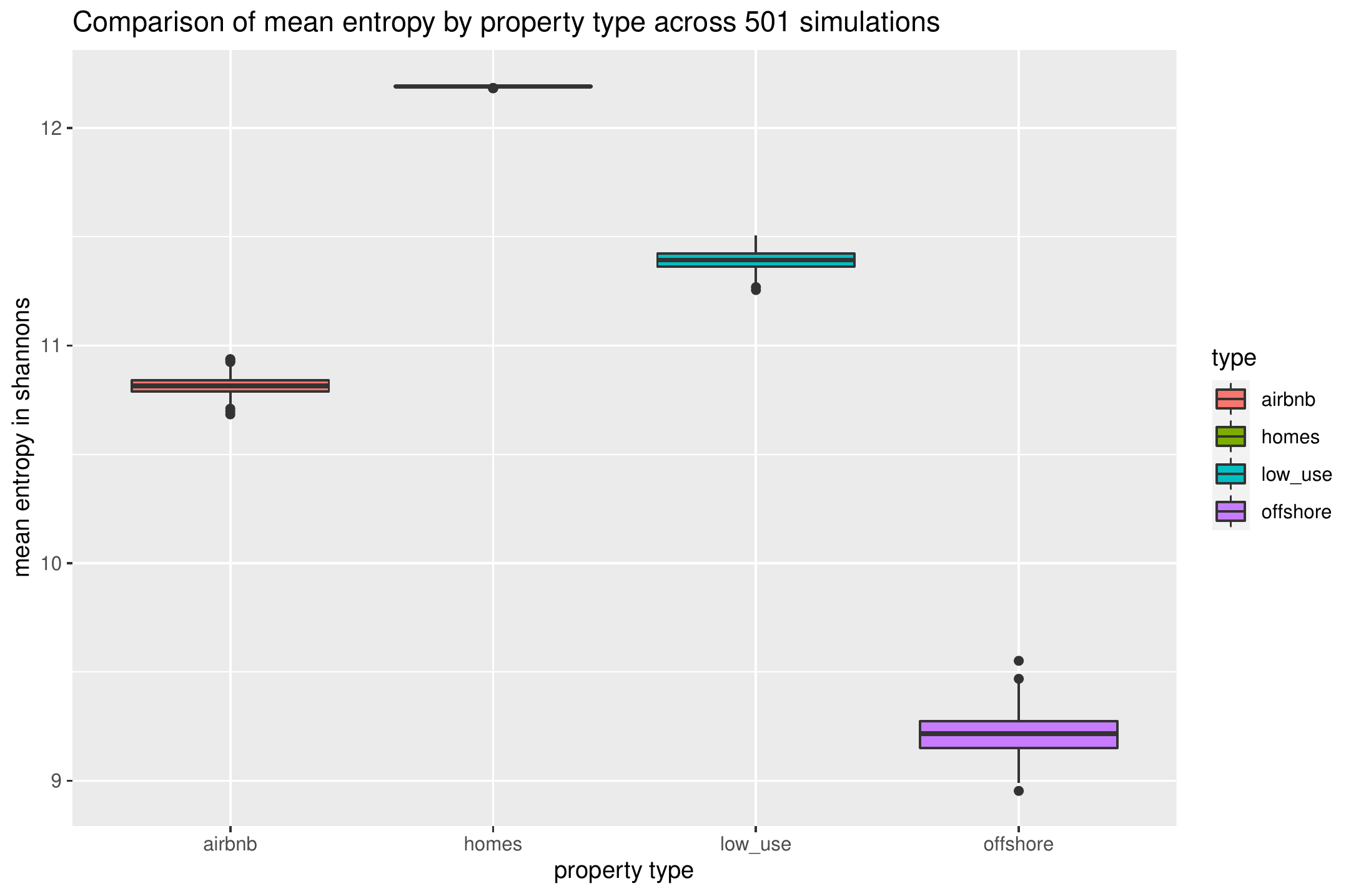}
    \caption{The entropy of the property types appears inversely proportional to the value. Indicating that the offshore, airbnb and low-use properties are not only more concentrated geographically that normal homes but are more concentrated into high value areas.}
    \label{fig:entropy_plot}
\end{figure}

Figure \ref{fig:property_maps} shows the distribution of the different property types across the capital. The map shows that offshore property is the most concentrated with a large fraction focused in the centre of the city. Airbnb properties are not as concentrated but appear to have high levels of spatial auto-correlation. In contrast the distribution of empty homes appears relatively widely dispersed and with lower levels of spatial auto correlation. The visual interpretation of Figure \ref{fig:property_maps} is supported by the quantative analysis using entropy and Moran's I. Figure \ref{fig:entropy_plot} shows that the entropy of the offshore property is substantially higher than that of the other property types. The mean entropy of the different property types is 9.21 for offshore, 10.8 for airbnb, 11.4 for low-use and 11.2 for all UDPs. This entropy ranking matches the mean value ranking of the property types, however this is not the case with the spatial auto-correlation. Table \ref{tab:metrics_table} shows that offshore property has a low spatial auto-correlation (0.47) compared to the other UDP types (airbnb 0.72, low-use 0.61). It should be noted that figures \ref{fig:total_value} and \ref{fig:mean_property_price} have very small variance, there is little doubt the differences between the categories is statistically significant.

\begin{table*}[ht]
\centering
\begin{tabular}{rlrrrrr}
  \hline
 & type & total value (£B) & counts (K) & mean value (£M) & bits & morans I \\ 
  \hline
1 & airbnb & 42.30 &  41 & 1.03 & 10.81 & 0.72 \\ 
  2 & low\_use & 85.10 &  93 & 0.91 & 11.39 & 0.61 \\ 
  3 & offshore & 56.10 &  42 & 1.33 & 9.21 & 0.47 \\ 
  4 & homes & 2287.70 & 3553 & 0.64 & 12.19 & 0.33 \\ 
  5 & UDP & 173.80 & 170 & 1.02 & 11.20 & 0.70 \\ 
   \hline
\end{tabular}
\caption{The complete metrics list for the analysis of the main property types} 
\label{tab:metrics_table}
\end{table*}

\subsubsection{Nested vs single offshore properties}

\begin{figure}
    \centering
    \includegraphics{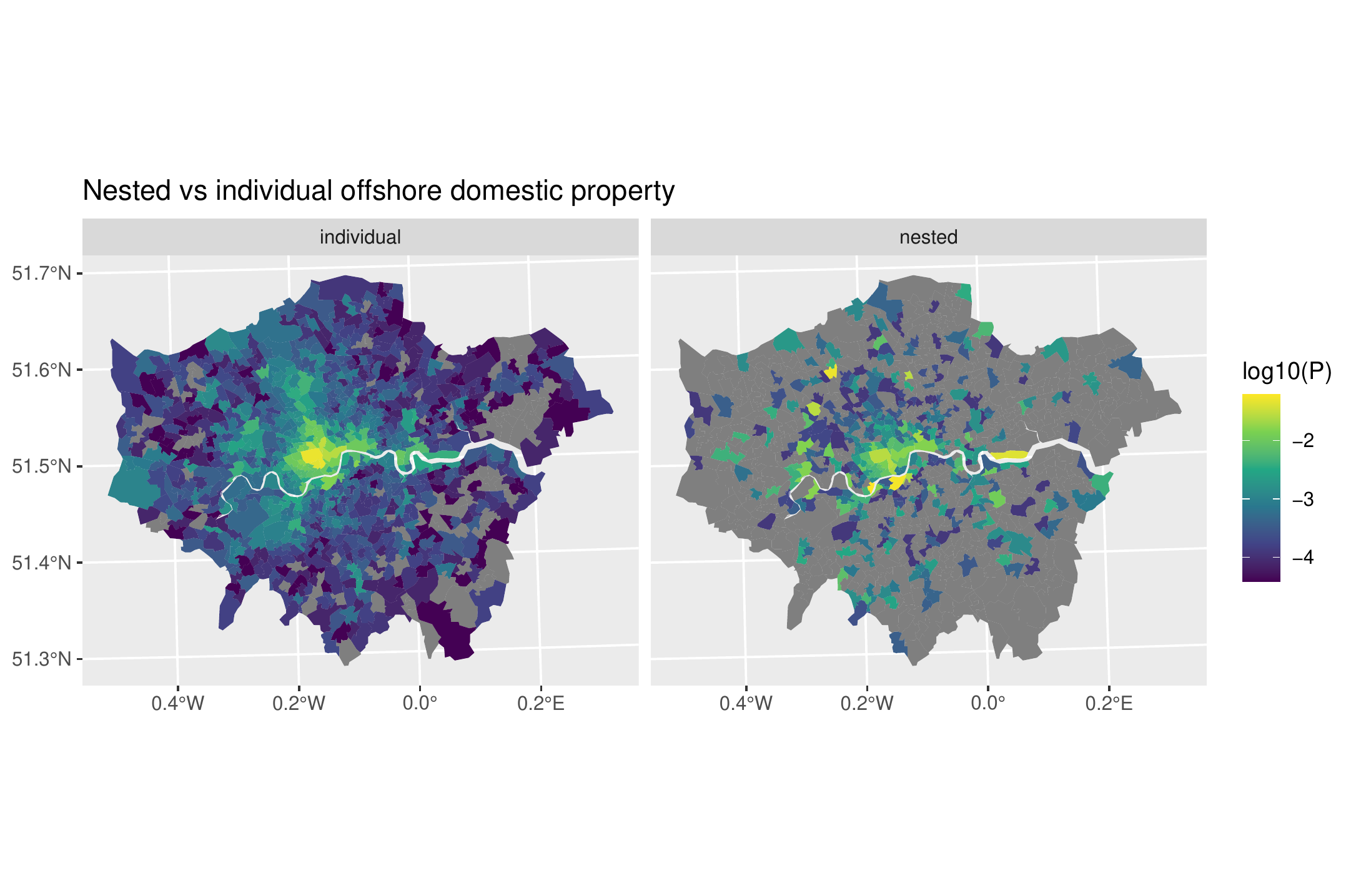}
    \caption{Comparing the nested vs normal property shows very different spatial patterns. These patterns correspond to large average differences in value. Grey areas have no properties of that type present}
    \label{fig:property_maps_nested}
\end{figure}

The notably low spatial auto-correlation value for offshore property compared to the other UDP types requires further investigation. One of the interesting aspects of the OCOD dataset is that it contains nested addresses. When the nested and single addresses are analysed separately a clear separate dynamic emerges. The single offshore properties show relatively high spatial auto-correlation with a value of 0.656 and a similarly low entropy as previously. However, the nested properties have a dramatically lower spatial auto correlation of 0.112 and a much lower entropy of 7.35 bits. There is also a substantial difference in the estimated values of these properties with the individual properties having a higher value of\pounds 1.5 Million whilst the nested properties are closer to the value of conventional housing with a mean value of \pounds 890 thousand. Table \ref{tab:metrics_table_nested}, provides the full comparison of this analysis, whilst figure \ref{fig:property_maps_nested}, maps the differences between the two types of offshore property. Looking at the price paid for individual properties provided in the original OCOD dataset, shows the mean to be \pounds 1.8 Million, removing all properties that are suspiciously low and the mean value increases to \pounds 3 Million. However, the prices in the OCOD dataset have a lot of quality issues and businesses that have been miss-classified will easily skew the total. Having said some properties within the dataset do have confirmed sales prices of over \pounds 100 million. Overall even though it may be slightly conservative the sampling method prevents unrealistically high estimates.

\begin{table*}[ht]
\centering
\begin{tabular}{rlrrrrr}
  \hline
 & type & total value (£B) & counts (K) & mean value (£M) & bits & morans I \\ 
  \hline
1 & individual & 77.60 &  51 & 1.51 & 9.55 & 0.66 \\ 
  2 & nested & 17.30 &  16 & 1.06 & 7.35 & 0.11 \\ 
   \hline
\end{tabular}
\caption{The complete metrics list comparing nested and single offshore homes} 
\label{tab:metrics_table_nested}
\end{table*}

\subsubsection{Country of Incorporation}

There are clear patterns in the London domestic property market with regards country of incorporation of offshore companies. Whilst there are properties associated with 133 different countries or territories, 50\% of the properties are related to companies incorporated in the three crown dependencies of Jersey, Guernsey and the Isle of Man. A further 30\% of properties are associated with one of The British overseas territories. In 
fact only 4 territories, British Virgin Islands, Jersey, Guernsey and The Isle of Man, are associated with 78\% of all properties. Looking at the relationship with nested properties, The crown dependencies have an over representation of nested properties with The crown dependencies and the BVI representing 74\% of all nested titles and 91\% of all nested properties. Jersey alone represents 52\% of all nested properties. More detail can be found in supplementary material.

\FloatBarrier
\section{Discussion}

The overall labelling quality was high, an f1 score of 0.96 suggests that the data is being effectively labelled and provides confidence in the quality of the final addresses. This also shows that the deduction used in the type two classifier is a robust approach. On top of this the final classification values also had strong performance across the categories although separating businesses and domestic properties was problematic at times. Improving the performance of the NER process would require a more in depth analysis of why buildings were not being captured at the same level as the other entities possibly including large scale visual data and as such is beyond the scope of this paper. Recall was notably lower for the buildings labels. This was partially due to the complexity of building names, but the high precision suggests that there may be some patterns which have not been properly addressed. A failure to identify evenly patterns from across all building names highlights a drawback of the weak learning approach and a risk of inducing systemic bias. It was surprising that the RoBERTa based spaCy model did not improve performance, this may be because the structure of the addresses is so simple and regular that the attention abilities of the transformer could not add any value. The lower recall and precision scores for Business and Domestic properties was mostly due to these two classes being difficult to separate. This challenge highlights the importance of the UPRN system in the UK. It should be noted that a freedom of information request was sent to both the VOA and the Land Registry requesting they provide the UPRN data, however, both FOI's were rejected.

Although undoubtedly the most appropriate choice for the OCOD dataset, the parser developed for this paper is a niche product designed for a single use case and is unlikely to offer the robustness or flexibility of  already available general purpose parsers such as Deepparse \cite{yassine_leveraging_2020} and libpostal \cite{barratine_libpostal_2017}. We would not recommend it as a first choice for parsing data other than the OCOD dataset and possibly other UK government address data. The code for creating the enchanced OCOD dataset is publicly available under a GNU GPLv3 license. In addition the supplementary material provides a more in depth description of the training details. 
Having the trained pipeline available means those who wish to use the enhanced OCOD dataset can do so for any past or future release of the Land Registry's OCOD dataset. 

The results of analysing the enhanced OCOD dataset show that there is a substantial amount of UDP in London (Property that is either low-use, Airbnb or offshore owned). There is an estimated \pounds 145-174 billion UDP in London. This is approximately 7.5\% of the total domestic value, of this there is approximately \pounds 56 Billion offshore owned property representing 42 thousand dwellings. We found that there were two distinct types of offshore property single and nested. The single properties are may be under valued as an analysis of the prices in the OCOD dataset suggest a mean value of \pounds 3 Million whilst the empirical sample only returns \pounds 1.5 Million. However, due to data quality concerns with the values reported in OCOD the mean reported value should only be considered as context. In the dataset nested properties, those properties sharing a title number, are mostly entire housing complexes and have very different spatial patterns to the individual offshore properties. Individual offshore properties are very expensive even by the standards of UDP, in addition they are concentrated on the centre of London with strong spatial auto correlation. In contrast nested offshore property is somewhat less concentrated on central London but more highly concentrated in general, there is also almost no spatial correlation. There was a brief analysis relating to country of incorporation. Four territories control 78\% of all offshore owned domestic property London. These territories are the three UK Crown Dependencies (Jersey, Guernsey, The Isle of Man) and The British Virgin Islands, which is a British Overseas Territory. Although we do not suggest that any specific property is being used to launder money, throughout the project certain properties did show some unusual characteristics; However a deeper inspection of these properties is beyond the scope of this paper.

\section{Conclusion}
This paper proposes a new method of identifying and classifying offshore owned domestic property in England and Wales. Using the dataset created by this method we have shown that Offshore owned domestic property in London is not a homogeneous group, but contains at least two distinct types, characterised by measurable quantities and appearing to relate to purpose and country of incorporation. We have also shown the distribution and value of three different types of UDP and that all three property types are more concentrated and more expensive than conventional domestic property. Whilst this paper does not draw any causal relations or make any policy proposals it provides for the first time a clear map to understanding the behaviours and value of Offshore owned domestic property and Low-use property in London. This data can be used by sociologists, economist and policy makers to ensure that attempts to reduce money laundering and high property prices are based on detailed data that reflect the real situation. In addition to the findings of the analysis the paper has created a high quality open-source dataset that allows further research as well as a code pipeline to create new datasets as required. It has been said that fighting money laundering requires openness and transparency \cite{mitchell_accountants_1998}, the OCOD dataset was a step in that direction, we hope that the Enhanced OCOD dataset is the next.

\subsection{Future work}
The enhanced OCOD dataset can be used for a wide variety of projects. Future work can expand the offshore analysis to the entire of England and Wales. Building on the work of \cite{sa_effect_2016}, and analysis of the nested offshore properties can be performed to explore whether they are vehicles for gentrification. Some of the properties appeared to have unusual characteristics, work of \cite{unger_money_2011} could be used to identify if these properties may be being used for money laundering.

\section{acknowledgements}
We would thank humanloop.com for giving us access to the `Programmatic' beta and providing technical support during the NER labelling phase of the project. 

\section{funding}
The authors would like to thank Trust for London (grant number MAIN-S2-06.10.2020-8792(6930) ) for funding the research. We would also like to thank Kingston University for providing additional research funding.

\bibliographystyle{unsrt}

\bibliography{bibliography.bib}

\end{document}